\documentclass{svproc}
\usepackage{graphicx}%
\usepackage{multirow}%
\usepackage{amsmath,amssymb,amsfonts}%
\usepackage{mathrsfs}%
\usepackage[title]{appendix}%
\usepackage{xcolor}%
\usepackage{textcomp}%
\usepackage{manyfoot}%
\usepackage{booktabs}%
\usepackage{algorithm}%
\usepackage{algorithmicx}%
\usepackage{algpseudocode}%
\usepackage{listings}%
\usepackage{url}

\begin{document}
\mainmatter              
\title{Local Intrinsic Dimensionality for Dynamic Graph Embeddings} 
\titlerunning{LID for Dynamic Graph Embeddings}  
%
\author{Du\v{s}ica Kne\v{z}evi\'{c} \and Milo\v{s} Savi\'{c} \and Milo\v{s} Radovanovi\'{c}}
\authorrunning{Du\v{s}ica Kne\v{z}evi\'{c} et al.} 

\institute{Department of Mathematics and Informatics, Faculty of Sciences,\\ University of Novi Sad, Serbia\\
Trg Dositeja Obradovi\'{c}a 3, 21000 Novi Sad\\
\email{\{lucy, svc, radacha\}@dmi.uns.ac.rs}}

\maketitle              

\begin{abstract}
The notion of local intrinsic dimensionality (LID) has important theoretical implications and practical applications in the fields of data mining and machine learning. Recent research efforts indicate that LID measures defined for graphs can improve graph representational learning methods based on random walks. In this paper, we discuss how NC-LID, a LID measure designed for static graphs, can be adapted for dynamic networks.  Focusing on dynnode2vec as the most representative dynamic graph embedding method based on random walks, we examine correlations between NC-LID and the intrinsic quality of 10 real-world dynamic network embeddings. The obtained results show that NC-LID can be used as a good indicator of nodes whose embedding vectors do not tend to preserve temporal graph structure well. Thus, our empirical findings constitute the first step towards LID-aware dynamic graph embedding methods.
\keywords{dynamic graphs, graph embeddings, local intrinsic dimensionality}
\end{abstract}

\section{Introduction}\label{intro_sec}

Many real-world complex systems studied in diverse research and engineering disciplines can be modeled by graphs or networks showing relations or interactions among their constituent units~\cite{Savic2019Fundametal}.  Such complex systems are typically dynamic, meaning that the structure of underlying networks evolves in time by adding and deleting both nodes and edges~\cite{HOLME2012}. Networks representing complex dynamical systems are also called {\it dynamic} or {\it temporal networks}.

Graph-structured data describing complex dynamical systems can be used to construct various types of useful predictive models relying on machine learning (ML) algorithms and methods~\cite{Barros2021,Khoshraftar2024}. Typical examples include models for node classification, node clustering, link prediction, node/link attribute prediction, anomaly detection, diffusion prediction, etc.  There are three different approaches to build such predictive models: (1) by graph-native ML techniques (e.g., iterative classification, community detection), (2) by applying ML algorithms designed for tabular data to graph embeddings, and (3) by graph neural networks~\cite{Tomcic2024}. 

The focus of this paper is on dynamic graph embedding algorithms. In contrast to graph neural networks showing superior performance when building task-specific predictive models for feature-rich attributed graphs, graph embedding algorithms provide a task-agnostic approach for learning general-purpose graph representations of feature-sparse and non-attributed networks. The most numerous, scalable and robust dynamic graph embedding methods are ones based on random walks~\cite{Barros2021,Khoshraftar2024}. The main idea of these methods is to sample a certain number of random walks for each node that are then treated as sentences over node identifiers. In this way, the problem of generating graph embeddings is reduced to the problem of generating text embeddings.

In our previous works~\cite{2021_NCLID_SISAP,SAVIC2023,Knezevic22}, it has been shown that static graph embedding methods based on random walks can be improved by taking into account local intrinsic dimensionality (LID)~\cite{Houle2013} aspects of nodes.  In~\cite{2021_NCLID_SISAP} we have proposed  a principal approach for formulating LID-related measures for nodes in static graphs and defined one concrete LID measure NC-LID that is based on natural communities~\cite{Lancichinetti_2009} as intrinsic node localities. Then, we have designed and evaluated two LID-aware node2vec~\cite{nodetovec2016} extensions based on the NC-LID measure showing that they improve intrinsic embedding quality~\cite{SAVIC2023}, as well as the performance of concrete machine learning applications~\cite{Knezevic22}.

In this work, we discuss how NC-LID can be adapted for dynamic networks (Section~\ref{LID_graphs}). As the main contribution, we propose a methodology to evaluate correlations between NC-LID and the intrinsic quality of dynamic graph embeddings (Section~\ref{evalmethodology}). The application of the proposed methodology to one concrete dynamic graph embedding method (dynnode2vec~\cite{dynnode2vec}) and various real-world dynamic networks reveal that NC-LID is able to point to nodes with poorly constructed embedding vectors that do not preserve local connectedness of the corresponding nodes (Section~\ref{results}). Therefore, just as for static graph embedding methods, these results indicate that appropriate LID-aware extensions, outlined in the last section of the paper (Section~\ref{conclusionfuture}), have the potential to improve dynamic graph embedding methods. 

\section{Local Intrinsic Dimensionality Measure for Dynamic Networks}\label{LID_graphs}

The intrinsic dimension (ID) $d'$ of a set of data points $D$ in a $d$-dimensional Euclidean space can be intuitively understood as the minimal number of features required to represent $D$ in a lower-dimensional space ($d' < d$) without significant information loss, or as the dimension of the surface achieving the best fit to $D$. However, ID may vary across a dataset leading to the notion of local intrinsic dimensionality (LID), with the main idea to asses ID of local data spaces surrounding individual data points. In his seminal paper, Houle~\cite{Houle2013} used the ball-expanding model to formally define LID w.r.t. the distribution of distances to a reference data point $p$. Intuitively, LID expresses the degree of difficulty to separate nearest neighbors of $p$ from the rest of the dataset, thus reflecting the degree of indiscriminability of the underlying distance function in the vicinity of~$p$.  
 
The works by Savi\'{c} et al.~\cite{2021_NCLID_SISAP,SAVIC2023} were the first attempts to define LID measures for nodes in a graph. Building on Houle's LID foundations for tabular datasets, Savi\'{c} et al. proposed an approach for formulating LID-related measures for graphs following the idea of substituting a ball around a data point with a subgraph around a node in order to be able to estimate the discriminatory power of a graph-based distance of interest. A measure called NC-LID was introduced for the shorest-path distance as the underlying distance function and natural or local communities~\cite{Lancichinetti_2009} of nodes as their intrinsic localities.  

A community in a graph is a highly cohesive and dense subgraph sparsely connected to the rest of the graph~\cite{Savic2019Fundametal}. The natural community of node $n$ is a community
recovered from $n$ as the seed node. NC-LID relies on natural communities identified by the fitness-based algorithm proposed by Lancichinetti et al.~\cite{Lancichinetti_2009}. Let $S$ be the natural  community of $n$ in an unweighted graph $G = (V, E)$, and let $k$ denote the largest shortest-path distance from $n$ to any node in $S$.  The NC-LID of $n$ is then formally defined as:
$$\mbox{NC-LID}(n) = - \ln \left( \frac{|S|}{D(n, k)} \right),$$
where $|S|$ is the number of nodes in $S$ and $D(n, k)$ is the number of  nodes whose shortest-path distance from $n$ is smaller than or equal to $k$~\cite{2021_NCLID_SISAP,SAVIC2023}. The lowest possible value of NC-LID is equal to 0. In such cases, the shortest-path distance can perfectly separate $S$ from the rest of the nodes in $G$, i.e., $S$ encompasses all nodes that are maximally $k$ hops away from $n$. Higher values of NC-LID($n$) imply that it is harder to distinguish $S$ from the rest of $G$ relying on the shortest-path distance. This means that with higher NC-LID natural communities take more complex, irregular shapes.  It is easy to see that NC-LID values of nodes in weighted graphs can be obtained by (1) using an appropriate distance measure instead of the shortest-path distance (e.g., Dijsktra's distance), and (2) recovering natural communities relying on the fitness function that uses node strength instead of node degree.

The NC-LID measure defined for static graphs can be adapted for dynamic graphs in two principal ways depending on the chosen temporal graph representation model. In discrete-time representation models, the evolution of a dynamic graph is described by a sequence of static graphs that are called snapshots.  Thus, for a node we have a sequence of NC-LID values considering those snapshots in which the node appears. Since dynamic graph algorithms designed for discrete-time representations incrementally operate on static graphs, the original NC-LID definition for static graphs can be used without any modifications when designing their LID-aware counterparts. 

The evolution of a graph in continuous-time representation models is given by a sequence of events that correspond to interactions between nodes. Events are usually represented by triplets in the form $(x, y, t)$, where $x$ and $y$ are two nodes involved in the interaction that occurs at time $t$. Additionally, the duration of interactions can be specified if interactions are not instantaneous.\footnote{It should be emphasized that in the most of publicly available datasets the duration of interactions is not provided.} Algorithms designed for continuous-time representation models usually operate on a single graph that aggregates all the events.  Edges in that aggregated graph are time-stamped, i.e., each edge is attributed with time points or time intervals at which the corresponding interaction occurs. In such setting, node distances can be defined considering temporal or time-varying paths~\cite{Kumar2011}. More specifically, for two events $(x, y, t_{1})$ and $(y, z, t_{2})$ there is a temporal path $x \rightarrow y \rightarrow z$  if $t_{2} > t_{1}$. The distance between two nodes $i$ and $j$ at time $t$ can be defined as the shortest time it takes to reach $j$ from $i$ at time $t$ along temporal paths. The temporal distance between $i$ and $j$ can then be obtained by averaging time-dependent distances over the entire period (from the first to the last event).  Consequently, the NC-LID of a node can be obtained by the static graph definition in which the shortest-path distance is replaced by the temporal distance and without any modification regarding natural community detection.  

\section{Evaluation Methodology}
\label{evalmethodology}

The aim of this work is to evaluate the ability of the NC-LID measure to improve dynamic graph embedding algorithms based on random walks. We consider discrete-time representations of dynamic graphs, since most of the existing dynamic graph embedding methods are based on this approach~\cite{Barros2021}.   

To obtain embeddings of real-world dynamic graphs we use dynnode2vec~\cite{dynnode2vec}. This method is an adaptation for dynamic graphs of the most representative random-walk based static graph embedding method, node2vec~\cite{nodetovec2016}. As the input dynnode2vec accepts an evolutionary sequence of graph snapshots. Similarly as node2vec, embedding construction is based on biased random walks, whose sampling is controlled by four parameters: the number of random walks sampled from each node, the length of random walks, the return parameter $p$ controlling the probability of going back to the previous node in the walk, and the in-out parameter $q$ controlling the probability of exiting the immediate neighborhood of the previous node. Using the first snapshot in the input sequence, dynnode2vec samples random walks to construct an initial skip-gram model, performing both steps in the same way as node2vec. The initial model is then incrementally updated with each subsequent snapshot by random walks sampled from the so-called evolutionary nodes, i.e., nodes whose neighborhoods have changed in two consecutive snapshots.  Node embedding vectors are constructed from the skip-gram model after each model update, which means that dynnode2vec provides embeddings for each individual snapshot in the evolutionary sequence.

The main goal of general-purpose graph embedding methods is to preserve structural properties of networks in embedded spaces. Thus, well performing graph embedding methods need to result with embeddings from which original graphs can be reconstructed~\cite{SAVIC2023}. The graph reconstruction process is performed by computing the Euclidean distance between embedding vectors for each pair of nodes and then connecting  the $|E|$ closest node pairs, where $|E|$ is the number of links in the original graph.  Let $n$ denote an arbitrary node and let $C$ be the number of correctly reconstructed links incident to $n$.  Then, the following embedding evaluation metrics can be computed:
\begin{itemize}
\item Precision -- $C$ divided by the number of links $n$ has in the reconstructed graph.
\item Recall -- $C$ divided by the number of links $n$ has in the original graph.
\item $F_{1}$ score -- the harmonic mean of precision and recall. 
\end{itemize}
Higher values of precision, recall and $F_{1}$ imply a lower number of link reconstruction errors for $n$.  Precision, recall and $F_{1}$ at the graph level can be obtained by micro-averaging over the nodes.

In our methodology, dynnode2vec is tuned by finding values of its hyper-parameters $p$ and $q$ that give embeddings maximizing the $F_{1}$ score in the last snapshot for five different graph embedding dimensions (10, 25, 50, 100 and 200).  For $p$ and $q$ we consider values in \{0.25, 0.50, 1, 2, 4\}, while the number of  random walks per node and the length of each random walk are fixed to 10 and 32,  respectively.  For one configuration of hyper-parameters, the $F_{1}$ score is determined by averaging the results of 10 runs.
 
To examine how NC-LID is related to the quality of dynamic graph embeddings, we compute Spearman correlations between NC-LID and above defined embedding evaluation metrics over all graph snapshots.  Significant negative correlations between NC-LID and embedding evaluation metrics indicate that NC-LID can point to nodes that have poorly constructed embedding vectors. Additionally, throughout the whole evolutionary sequence, nodes are assigned to two categories: 
\begin{enumerate}
	\item nodes having NC-LID above the average NC-LID in the current snapshot (denoted by $H$), and 
	\item nodes having NC-LID below the average NC-LID in the current snapshot (denoted by $L$).
\end{enumerate}
Then,$F_{1}$ scores of those two groups of nodes are compared by the Mann-Whitney U (MWU) test~\cite{MWU1947}. MWU is a test of stochastic equality checking the null hypothesis that $F_{1}$ scores in $H$ do not tend to be neither greater nor smaller than $F_{1}$ scores in $L$.  Additionally, we compute two probabilities of superiority reflecting the strength of stochastic inequality: 
\begin{enumerate}
	\item PS($H$) -- the probability that a randomly selected node from $H$ has a higher $F_{1}$ than a randomly selected node form $L$,
	\item PS($L$) - the probability that a randomly selected node from $L$ has a higher $F_{1}$ than a randomly selected node from $H$.
\end{enumerate}

Additionally, we investigate whether high NC-LID nodes tend to be the most structurally important nodes in the network. To address this question, we compute the Spreman correlation coefficient between NC-LID and widely used node centrality metrics:  degree, betweenness, closeness, eigenvector and shell index centrality~\cite{Savic2019Fundametal}.   

\section{Results and Discussion}
\label{results}

Our experimental evaluation of NC-LID is conducted on real-world dynamic networks listed in Table~\ref{table_net_chars}. Examined networks are typically used dynamic networks in the literature and can be retrieved from two publicly-available repositories: SNAP\footnote{https://snap.stanford.edu/data/\#temporal} and Network Repository.\footnote{https://networkrepository.com/dynamic.php} Besides the total number of nodes (denoted by $|V|$) and the total number of links (denoted by $|E|$) that can be observed throughout network evolution, the number of snapshots and time resolution of each snapshot,  Table~\ref{table_net_chars} also shows the average activation of nodes and links, denoted by $a(V)$ and $a(E)$, respectively. Namely, the activation of a node (resp. link) is the number of snapshots in which the node (resp. link) is present or active. 

\begin{table}[htb!]
\caption{Experimental datasets.}
\begin{center}
\addtolength{\tabcolsep}{3.5pt}
\begin{tabular}{@{}lllllll@{}}
\noalign{\smallskip}\hline \noalign{\smallskip}
Network & $|V|$ & $|E|$ & snapshots & time resolution & $a(V)$ & $a(E)$ \\
\noalign{\smallskip}\hline \noalign{\smallskip}
ia-hospital & 75 & 1369 & 4 & day & 2.79 & 1.34 \\
ia-contacts & 113 & 2470 & 3 & day & 2.65 & 1.19 \\
ia-enron & 151 & 2047 & 11 & 3-months & 6.57 & 1.74 \\
radoslaw-email & 167 & 4519 & 9 & month & 7.5 & 2.35 \\
ia-primschool & 242 & 9843 & 2 & day & 1.98 & 1.16 \\
fb-forum & 898 & 8212 & 5 & month & 3.03 & 1.28 \\
email-eu & 986 & 22205 & 17 & month & 12.29 & 2.79 \\
college-msg & 1894 & 14474 & 6 & month & 2.23 & 1.08 \\
fb-messages & 1899 & 16415 & 7 & month & 2.45 & 1.07 \\
ia-realitycall & 6763 & 7947 & 15 & week & 2.24 & 2 \\
\noalign{\smallskip}\hline \noalign{\smallskip}
\end{tabular}
\label{table_net_chars}
\end{center}
\end{table}

\begin{figure}[htb!]
	\centering
	\includegraphics[scale=0.57]{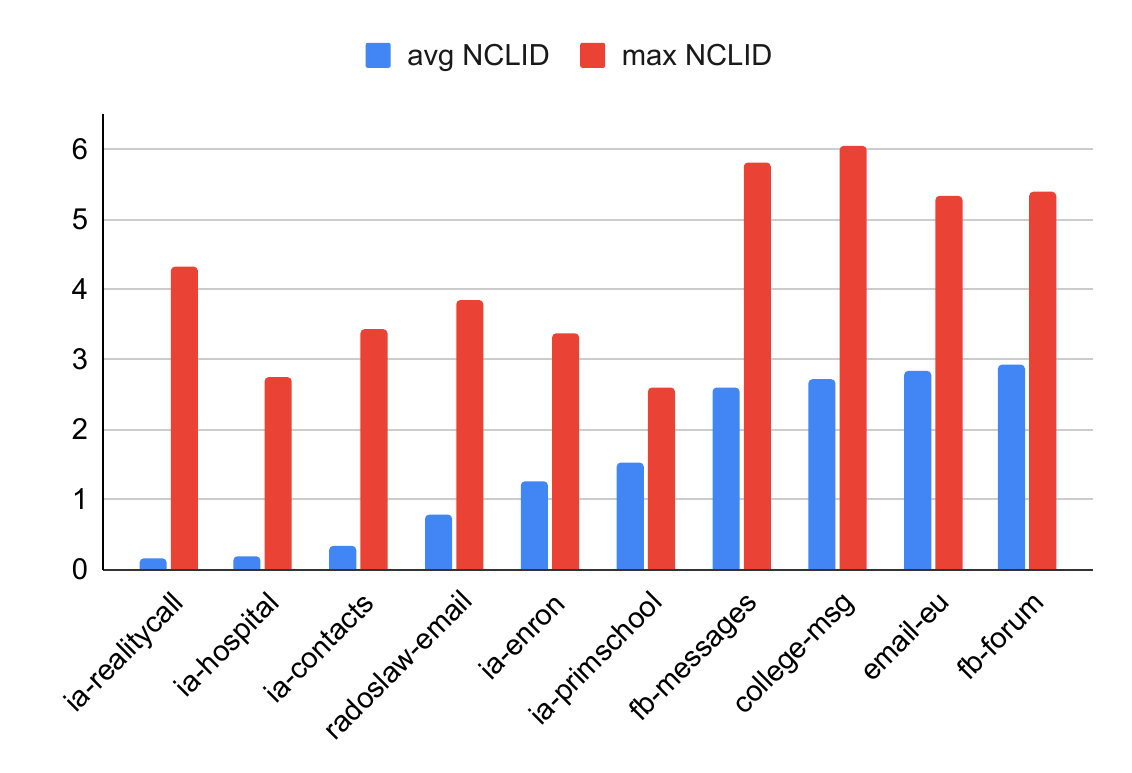}
	\caption{Average and maximal NC-LID per network.}
	\label{fig_NCLID}
\end{figure}

The average and maximal NC-LID over all snapshots for examined dynamic graphs are shown in Figure~\ref{fig_NCLID}, sorted from the lowest to the highest average NC-LID.  It could be observed that the average NC-LID does not depend on network size and density, i.e., smaller (resp., sparser) networks may have more complex natural communities than larger (resp., denser) networks.  For example, the largest average NC-LID is exibited by \texttt{fbforum}, which is 6th (out of 10 networks) by size and 4th by density.

\begin{figure}[htb!]
	\centering
	\includegraphics[scale=0.43]{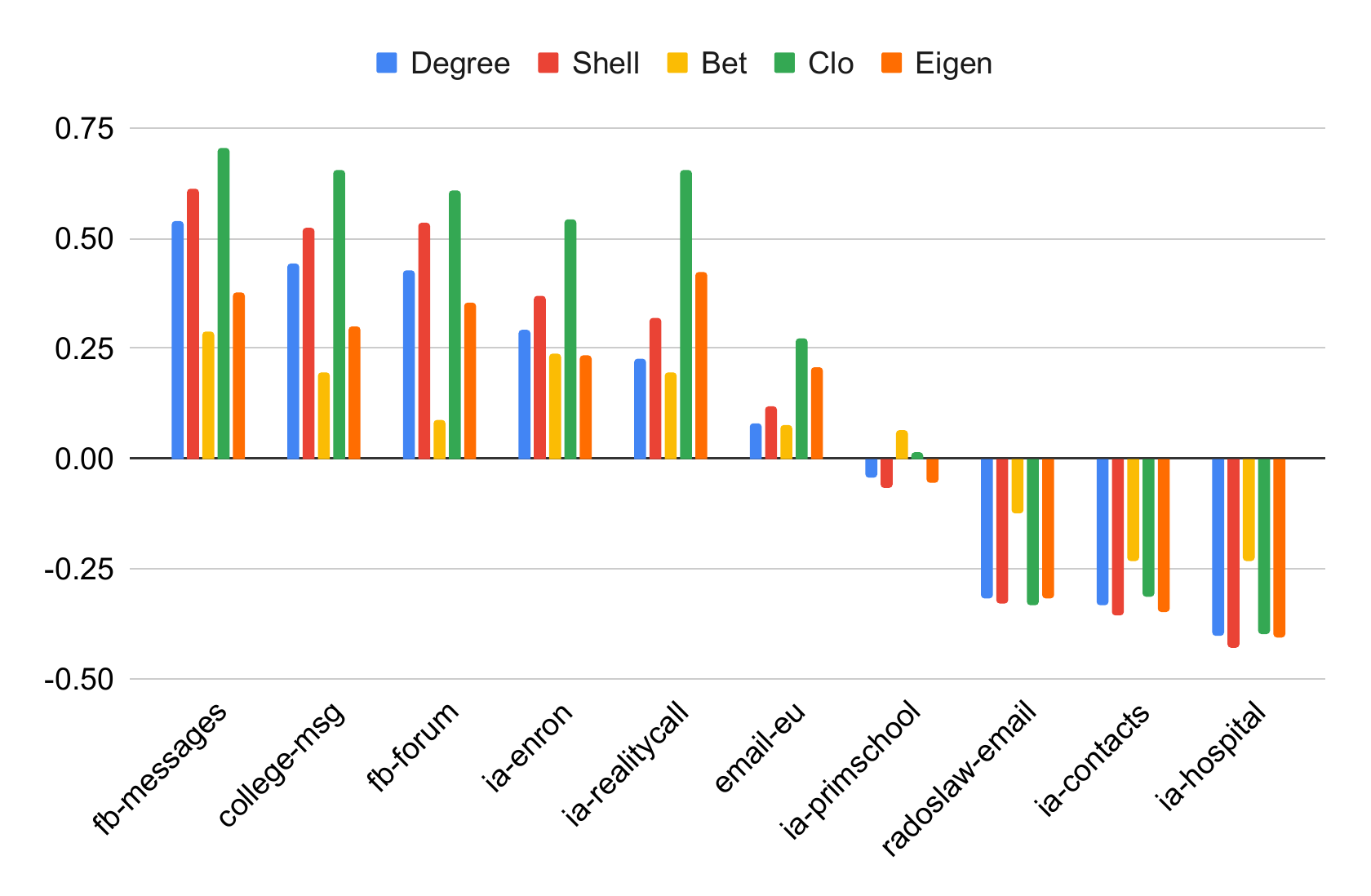}
	\caption{Spearman correlations between NC-LID and node centrality measures.}
	\label{fig_correlations_NCLID_centrality}
\end{figure}

Figure~\ref{fig_correlations_NCLID_centrality} shows Sperman correlations between NC-LID and node centrality metrics.  It can be seen that in 5 networks there are significant positive correlations, in 3 networks significant negative correlations, while in 1 network correlations are not strong. The strongest correlations, either positive or negative, are present for closeness centrality. This implies that, in some networks, nodes with complex natural communities tend to be located at network periphery (negative correlations), while in some networks such nodes tend to occupy central positions and be the most important nodes for dynamic processes running on networks.

The results of dynnode2vec tuning are presented in Table~\ref{table_n2v_bestembs}. 
This table shows the maximal $F_{1}$ scores for dynamic networks from our experimental corpus sorted 
from the highest to the lowest $F_{1}$, the dimension in which the maximal $F_{1}$ score 
is achieved (Dim.), precision, recall and the corresponding values of $p$ and $q$.
It can be observed that for 9 dynamic networks (all except \texttt{ia-realitycall}) we have $F_{1}$ scores in the range $[0.56, 0.83]$. Such high $F_{1}$ scores imply that dynnode2vec embeddings preserve the structure of input 
dynamic networks to a great extent.  Additionally, it can be seen that precision and recall tend to have close values implying that the corresponding quality attributes tend to be balanced in the best dynnode2vec embeddings. Consequently, it can be concluded that dynnode2vec achieves a very good performance at making general-purpose vector representations of nodes in dynamic networks, justifying our choice of this particular state-of-the-art dynamic graph embedding method for the purpose of NC-LID evaluation.

\begin{table}[htb!]
\caption{Characteristics of the best dynnode2vec embeddings (sorted from the highest to the lowest $F_{1}$ score).}
\begin{center}
\addtolength{\tabcolsep}{5pt}
\begin{tabular}{@{}lllllll@{}}
\noalign{\smallskip}\hline \noalign{\smallskip}
Graph & Dim. & $p$ & $q$ & Precision & Recall & $F_{1}$ \\
\noalign{\smallskip}\hline \noalign{\smallskip}
fb-forum & 200 & 1 & 1 & 0.8289 & 0.8346 & 0.8317 \\
ia-primschool & 10 & 1 & 4 & 0.8051 & 0.7869 & 0.7959 \\
college-msg & 50 & 1 & 1 & 0.664 & 0.7043 & 0.6835 \\
email-eu & 50 & 0.5 & 0.5 & 0.6325 & 0.7174 & 0.6722 \\
ia-contacts & 200 & 0.5 & 4 & 0.6944 & 0.6158 & 0.6527 \\
ia-hospital & 100 & 0.5 & 4 & 0.6476 & 0.6087 & 0.6274 \\
radoslaw-email & 25 & 0.5 & 4 & 0.6662 & 0.5925 & 0.6271 \\
ia-enron & 25 & 0.5 & 2 & 0.4643 & 0.7221 & 0.5651 \\
fb-messages & 100 & 0.5 & 1 & 0.5146 & 0.6072 & 0.5569 \\
ia-realitycall & 10 & 4 & 0.5 & 0.0329 & 0.0404 & 0.0362 \\
\noalign{\smallskip}\hline \noalign{\smallskip}
\end{tabular}
\label{table_n2v_bestembs}
\end{center}
\end{table}

\begin{figure}[htb!]
\centering
\includegraphics[scale=0.57]{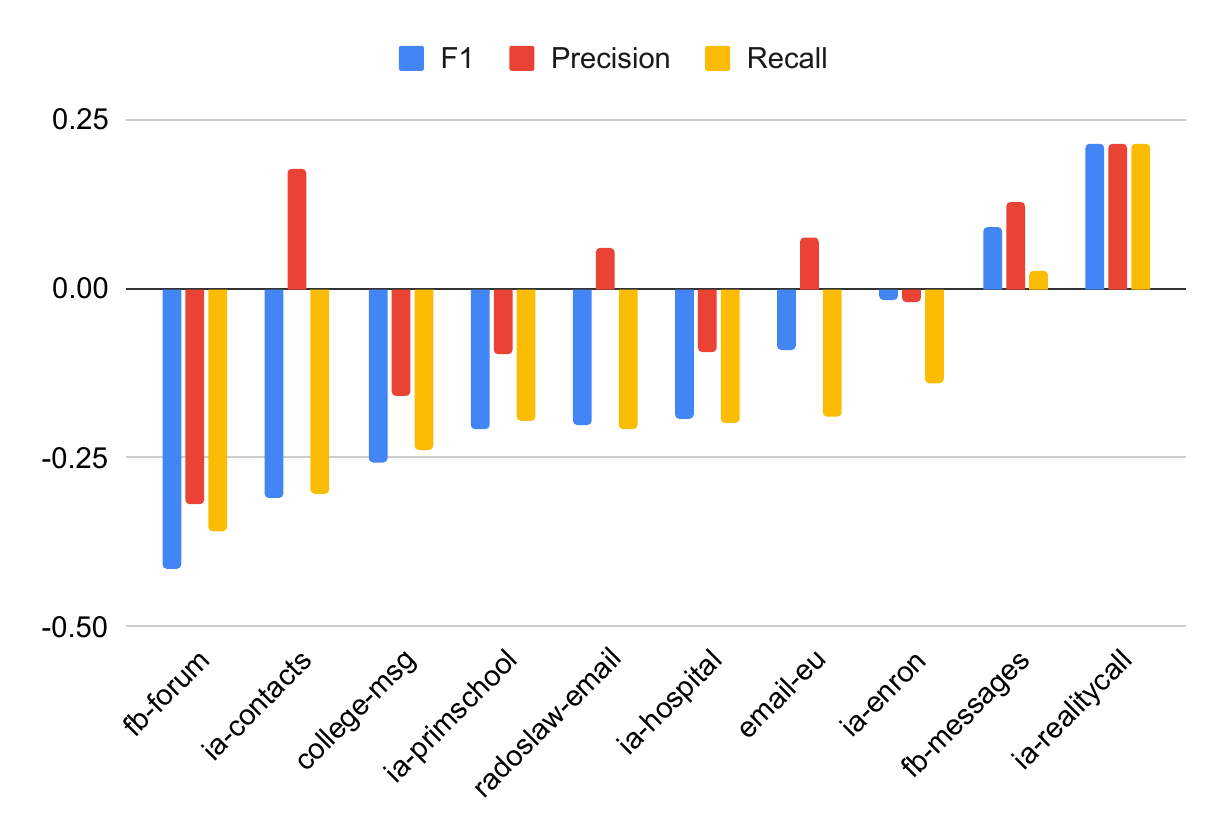}
\caption{Spearman correlations between NC-LID and graph embedding evaluation measures ($F_{1}$, precision and recall). The networks are sorted from the lowest to the highest correlation between NC-LID and $F_{1}$.}
\label{fig_correlations_NCLID_F1}
\end{figure}

High NC-LID nodes have more complex natural communities than low NC-LID nodes. Thus, it is quite reasonable to expect that high NC-LID nodes will have a higher number of link reconstruction errors due to more complex natural communities. To verify this assumption, we compute Spearman correlations between NC-LID and embedding evaluation metrics (precision, recall, $F_{1}$) obtained from the best dynnode2vec embeddings. Since lower values of precision, recall and $F_{1}$ imply a higher number of incorrectly reconstructed links, counting both false positives and false negatives, significant negative correlations between NC-LID and those embedding quality metrics would imply that NC-LID is able to point to nodes with `weak` representations (`weak` in the sense that their embedding vectors do not preserve their connectedness in the embedding space). The values of Spearman correlations are shown in Figure~\ref{fig_correlations_NCLID_F1}. It can be observed that for 8 networks there are negative Sperman correlations between NC-LID and $F_{1}$. Only for 2 networks we have positive correlations, where one of the networks is \texttt{ia-realitycall} (please recall that the dynnode2vec embedding for this particular network has extremely low precision, recall and $F_{1}$ scores, thus the correlation results obtained for this network can not be used to derive a meaningful conclusion regarding the relationship between NC-LID and embedding quality). 

 For the first 6 networks shown in Figure~\ref{fig_correlations_NCLID_F1} there are significant negative correlations between NC-LID and $F_{1}$ ranging from -0.19 to -0.41. For the same networks significant negative correlations between NC-LID and recall are also present (ranging from -0.19 to -0.36), and they also tend to exhibit significant negative correlations between NC-LID and precision with two exceptions (\texttt{ia-contacts} with a significant positive correlation and \texttt{radoslaw-email} without any significant correlation regarding precision). For the next two networks depicted in Figure~\ref{fig_correlations_NCLID_F1} we have significant negative correlations between NC-LID and recall. Therefore, it can be concluded that for the majority of networks in our experimental corpus there are notable negative correlations between NC-LID and at least one metric reflecting intrinsic embedding quality.

Table~\ref{table_mwu} summarizes the results of MWU statistical tests with accompanying probabilities of superiority.  $F_{1}(H)$ and $F_{1}(L)$ denote the average $F_{1}$ score for high NC-LID nodes ($H$) and low NC-LID nodes ($L$), respectively.  $U$ is the value of the MWU test statistic and the null hypothesis of the test is accepted if the $p$-value of $U$ is higher than 0.05, which is also indicated by the ACC column in Table~\ref{table_mwu} (``no'' means that the null hypothesis is rejected implying statistically significant differences between high and low NC-LID nodes w.r.t their $F_{1}$ scores). 

\begin{table}[htb!]
\caption{Comparison of $F_{1}$ scores of high NC-LID nodes ($H$) and
low NC-LID nodes ($L$) using the Mann-Whitney U test.}
\begin{center}
\addtolength{\tabcolsep}{3.5pt}
\begin{tabular}{llllllll}
\noalign{\smallskip}\hline \noalign{\smallskip}
Network & $F_{1}(H)$ & $F_{1}(L)$ & $U$ & $p$ & ACC & PS($H$) & PS($L$) \\
\noalign{\smallskip}\hline \noalign{\smallskip}
ia-realitycall & 0.0863 & 0.0393 & 17873699 & 7.13E-43 & no & 0.1701 & 0.0817 \\
college-msg & 0.4761 & 0.5817 & 1485719.5 & 5.75E-30 & no & 0.3692 & 0.5804 \\
email-eu & 0.5998 & 0.6253 & 13980209.5 & 1.52E-13 & no & 0.4495 & 0.5308 \\
radoslaw-email & 0.4863 & 0.5729 & 127076.5 & 7.92E-10 & no & 0.3824 & 0.6022 \\
ia-contacts & 0.3482 & 0.5517 & 2279 & 3.13E-08 & no & 0.2272 & 0.7678 \\
fb-forum & 0.5869 & 0.6487 & 697028.5 & 4.81E-08 & no & 0.4032 & 0.5296 \\
fb-messages & 0.575 & 0.4836 & 1436845.5 & 1.46E-07 & no & 0.5194 & 0.4125 \\
ia-hospital & 0.4003 & 0.5499 & 908.5 & 0.0015 & no & 0.2882 & 0.7031 \\
ia-primschool & 0.7544 & 0.7629 & 26816 & 0.2448 & yes & 0.4688 & 0.5303 \\
ia-enron & 0.5256 & 0.5309 & 57862.5 & 0.7213 & yes & 0.4791 & 0.4949 \\
\noalign{\smallskip}\hline \noalign{\smallskip}
\end{tabular}
\label{table_mwu}
\end{center}
\end{table}

The results of MWU tests reveal that there are statistically significant differences in $F_{1}$ scores between high NC-LID and low NC-LID nodes in 8 networks (the null hypothesis of the test is accepted for \texttt{ia-primschool} and \texttt{ia-enron}). Considering those 8 networks, in 6 of them we have that $F_{1}(H) < F_{1}(L)$ and $\mbox{PS}(H) \ll \mbox{PS}(L)$ implying that $F_{1}$ scores of high NC-LID nodes tend to be significantly lower than $F_{1}$ scores of low NC-LID nodes. 

\section{Conclusions and Future Work}
\label{conclusionfuture}

In this paper we discussed how NC-LID, a LID measure originally designed for static graphs, can be adjusted for dynamic graphs, considering both their discrete-time and continuous-time representations.  Then, following our previous works focused on static graphs, we proposed a methodology to examine the ability of NC-LID to point to weak parts of dynamic graph embeddings, i.e., to nodes whose embedding vectors do not tend to preserve their temporal neighborhoods. The obtained experimental results for the dynnode2vec dynamic graph embedding method on 10 real-world dynamic graphs indicate that NC-LID tend to negatively correlate to measures of intrinsic embedding quality obtained by the comparison of original graphs and graphs reconstructed from formed embeddings.

Our main finding suggests that NC-LID could be used to design LID-aware dynamic graph embedding algorithms. In our future work, inspired by our previous LID-elastic modifications of node2vec~\cite{SAVIC2023}, we will examine similar modifications for dynnode2vec and other dynamic graph embedding methods based on random walks. Since high NC-LID nodes tend to have low $F_{1}$ scores due to more complex intrinsic localities, their embedding vectors could be improved by personalizing hyper-parameters controling random walk sampling strategies based on NC-LID values.  For example, the number of random walks per node can vary so that a proportionally higher number of random walks is sampled for high NC-LID nodes, while keeping the total computational budget (the total number of random walk steps) constant.  Second, hyper-parameters controlling random walk biases (transition probabilities) could be also adjusted according to NC-LID values in order to prevent prematurely leaving complex intrinsic localities.

\subsubsection{Acknowledgments.}
This research is supported by the Science Fund of the Republic of Serbia, \#7462, Graphs in Space and Time: Graph Embeddings for Machine Learning in Complex Dynamical Systems -- TIGRA.

%
%
%

\bibliographystyle{spmpsci} 
\bibliography{refs} 
\end{document}